\documentclass{article}

\usepackage[main, final]{neurips_2025}

\usepackage[utf8]{inputenc} % allow utf-8 input
\usepackage[T1]{fontenc}    % use 8-bit T1 fonts
\usepackage[
pdftitle={NQS Scaling Law},
pdfcreator={pdflatex},
pdfsubject={preprint},
hyperindex = {true},
colorlinks = {true},
linkcolor = {blue},
citecolor = {blue}
]{hyperref}       % hyperlinks
\usepackage{url}            % simple URL typesetting
\usepackage{booktabs}       % professional-quality tables
\usepackage{amsfonts}       % blackboard math symbols
\usepackage{nicefrac}       % compact symbols for 1/2, etc.
\usepackage{microtype}      % microtypography
\usepackage{xcolor}         % colors

\usepackage{amsmath}
\usepackage{amssymb}

\usepackage{booktabs}
\usepackage{wrapfig}
\usepackage{amssymb,euscript,latexsym,amsmath}
\usepackage{xcolor}
\usepackage{graphicx} 
\usepackage{epstopdf}
\usepackage{asymptote}
\usepackage{bm} 
\usepackage{enumerate} 
\usepackage{color} 
\usepackage{setspace} 
\usepackage{subcaption} 
\usepackage{array}
\usepackage{cancel}
\usepackage{stmaryrd}

\usepackage[dvips]{epsfig}
\usepackage{epstopdf}
\usepackage{setspace}
\usepackage{lipsum}
\usepackage[shortlabels]{enumitem}
\usepackage{algorithm, algpseudocode}

\usepackage{cleveref}
\crefformat{equation}{(#2#1#3)}
\usepackage[affil-it]{authblk}

\newcommand{\abs}[1]{\left\lvert #1 \right\rvert}

\newcommand{\ip}[1]{\left\langle #1 \right\rangle}
\newcommand{\paren}[1]{\left( #1 \right)}
\newcommand{\braces}[1]{\left\{ #1 \right\}}
\newcommand{\bracket}[1]{\left[ #1 \right]}
\newcommand{\bra}[1]{\left\langle #1 \right\rvert}
\newcommand{\ket}[1]{\left\lvert #1 \right\rangle}

\newcommand\numberthis{\addtocounter{equation}{1}\tag{\theequation}}

\title{Large Language Model Scaling Laws for Neural Quantum States in Quantum Chemistry}

\author{%
  Oliver Knitter$^{1,2,*}$\quad
  Dan Zhao$^{2,3}$\quad
  \textbf{Stefan Leichenauer$^{2}$\quad
  Shravan Veerapaneni$^{1}$}\\
  \vspace{1em}
  $^1$University of Michigan\quad
  $^2$SandboxAQ\quad
  $^3$New York University\\
  {\normalfont *}\texttt{knitter@umich.edu}{\normalfont, currently at IonQ}
}

\affil{}

\begin{document}

\maketitle

\begin{abstract}
    Scaling laws have been used to describe how large language model (LLM) performance scales with model size, training data size, or amount of computational resources. Motivated by the fact that neural quantum states (NQS) has increasingly adopted LLM-based components, we seek to understand NQS scaling laws, thereby shedding light on the scalability and optimal performance--resource trade-offs of NQS ansatze. In particular, we identify scaling laws that predict the performance, as measured by absolute error and V-score, for transformer-based NQS as a function of problem size in second-quantized quantum chemistry applications. By performing analogous compute-constrained optimization of the obtained parametric curves, we find that the relationship between model size and training time is highly dependent on loss metric and ansatz, and does not follow the approximately linear relationship found for language models.
\end{abstract}

%\begin{abstract}
%    Scaling laws, often expressed as power laws, describe how large language model (LLM) performance scales with the number of parameters of the underlying model, training data size, or computational resources as they increase in size. As neural quantum states (NQS), a quantum-inspired classical analog to the variational quantum eigensolver (VQE) that uses deep neural networks to represent trial wavefunctions of a multi-qubit system, have begun using transformers and transformer-adjacent language model components, characterizing their scaling laws can shed light on their scalability and optimal performance-resource trade-offs. To assess the performance of NQS across different Hamiltonians of different problems in quantum chemistry, we use both absolute error and the V-score \cite{wu2024variational} to identify scaling laws for three distinct NQS ansatze for electronic ground structure calculations of second quantized molecular Hamiltonians: masked feedforward networks (MADE), transformers (NNQS--Transformer), and retentive networks (RetNets). \textcolor{red}{GIVE AN INTERESTING FINDING OR TWO FROM SCALING LAW BEHAVIOR}
%\end{abstract}

\section{Introduction}
\label{sec:intro}

Electronic ground state calculations are a fundamental problem in \textit{ab initio} quantum chemistry. As explicit calculation of the ground truth scales exponentially with molecule size, heuristic algorithms based on the variational principle have been devised including Hartree-Fock, coupled cluster methods \cite{bartlett2007coupled}, and tensor network methods such as DMRG \cite{schollwock2005density}. 

The incorporation of LLM components into NQS opens these ansatze up to systematic scalability analysis using scaling laws \cite{kaplan2020scaling}, mathematical relationships that describe how the computational requirements of a neural network model scale with the size of the underlying problem. For NLP, scaling laws have helped reveal how the performance of specific models grows with access to greater training data and compute resources. These laws help predict the limits of model capabilities, optimize the allocation of resources, and guide development of new architectures. 

Applying this broad analysis of NQS performance across different problem types is largely made possible by the introduction of V-Score, \cite{wu2024variational}, a problem-agnostic and model-agnostic metric analyzing performance of any algorithm designed to find the minimal eigenstate of a Hamiltonian. With it, the application of these scaling laws to NQS ansatze may prove fruitful, allowing us to systematically assess the efficiency and effectiveness of different neural architectures for representing quantum states within an application of such practical interest. Understanding the scaling behavior of autoregressive NQS ansatze is particularly important as it would:

\begin{center}
%\begin{itemize}[left=2em, right=3em]
\begin{itemize}
    \item Identify optimal neural architectures balancing accuracy, computational cost, and scalability. This is essential in quantum chemistry, where solving even fundamental problems requires ansatze that are sufficiently expressive at favorable model sizes. 
    \item Illuminate the strengths and weaknesses of NQS as an overall paradigm, highlighting specific areas where improvements in model design could lead to substantial gains in performance. 
    \item Gauge the broader applicability of these models for addressing larger molecular systems, more complex chemical environments, or general tasks beyond ground state energies.
\end{itemize}
\end{center}

% \subsection{Contributions}
\paragraph{Contributions.} We estimate and analyze Chinchilla-like scaling laws \cite{chinchilla} for autoregressive NQS models in solving/estimating electronic ground states. Drawing parallels with LLM scaling laws, we examine whether similar principles govern the efficiency and accuracy of deep learning architectures in the NQS domain. We focus on analyzing three specific autoregressive ansatze, based on three well-known architectures: MADE \cite{zhao2023nnqsmade}, transformers \cite{wu2023nnqstransformer}, and retentive networks (RetNets) \cite{knitter2024retnet_nqs}. 

Our analysis focuses on comparing the performance and costs of these architectures to better understand how they scale with model size and training time, relative to problem size. Inspired by similar estimation frameworks for LLMs \cite{kaplan2020scaling}, we introduce general floating point operation (FLOP) estimates for NQS problems as a proxy for compute costs. Our goal is to provide guidance for optimizing NQS performance in quantum chemistry by appropriately prioritizing the use of available compute resources for a given problem and molecule size.

Our paper is organized as follows. Section \ref{sec:background} gives an overview of the general problem formulation. Section \ref{sec:background_nqs} describes the NQS paradigm and the autoregressive ansatze under consideration. We detail the use of scaling laws in LLMs and how we apply them in our ansatz experiments for NQS in \cref{sec:background_scaling_laws}. We summarize  our findings in \cref{sec:result} and conclude with a discussion of our findings and future directions in \cref{sec:conclusion}.

\section{Related Work}
\label{sec:related_work}

Recent advances have used neural networks trained via variational Monte Carlo \cite{carleo2017solving} as an alternative variational ansatz. Known as neural quantum states (NQS), this method offers a potentially scalable solver for electronic ground state calculations \cite{Choo_2020} with many variants based on autoregressive ansatze \cite{barrett2022autoregressive, bennewitz2022neural, zhao2023nnqsmade} such as the  transformer architecture \cite{wu2023nnqstransformer} among others \cite{knitter2024retnet_nqs} as autoregressive ansatze, given the inherent expressiveness and utility of transformers within large language models (LLMs). However, further exploration is still necessary to understand how well these architectures perform and scale as NQS ansatze.

Scaling law analyses \cite{scaling_transfer, kaplan2020scaling} empirically study how LLM performance can scale with model or data size, computational resources, etc. The eponymous Chinchilla scaling laws \cite{chinchilla} estimated that doubling the size of a LLM requires a doubling in the number of training tokens for optimality. Since then, similar efforts have studied scaling behavior in transfer learning settings for downstream tasks \cite{scaling_downstream}, the influence on scaling behavior of inductive biases in model architectures \cite{scaling_arch}, effects of scaling on smaller language model trade-offs \cite{mini_scaling}, the potential impact of synthetic data on scaling and model collapse \cite{scaling_collapse}, how data-constrained language models scale with repeated data \cite{scaling_dataconstrain}, and scaling-informed optimization of model/data sizes when accounting for inference demand \cite{beyond_chinchilla}.

Scaling properties of NQS have been studied in time-evolved quantum states \cite{lin2022scaling} focusing on training unnormalized RBM ansatze \cite{carleo2017solving} and variations of the autoregressive NAQS ansatz \cite{sharir2020deep} under different Ising Hamiltonians. 
% This experiment directly compared the expressiveness of NQS with Matrix Product States (MPS), a one-dimensional tensor network method; results indicated that, similar to MPS, the number of NQS ansatz parameters needed to accurately learn these states grows exponentially with the amount of time they have been evolved. 
Our analysis is fundamentally similar with a few distinct caveats. We consider more sophisticated ansatz architectures, designed for greater expressiveness with respect to size. Furthermore, we are interested in exploring the scaling properties of autoregressive ansatze applied to a specific practical use case. While \cite{lin2022scaling} indicates NQS ansatze likely require exponentially many parameters to model all possible states of an exponentially large quantum system, the required expressiveness for practical applications may be much lower.

\section{Problem}
\label{sec:background}

\subsection{Preliminaries}

Calculating the energy of a molecule requires knowledge of its wavefunction, represented by a unit vector $\ket{\psi}$ in a complex Hilbert space, and its electronic Hamiltonian: a Hermitian operator $H_\text{e}$ on this space encoding the total kinetic and potential energy of the molecule. Its ground state energy corresponds to the minimal spectral value of $H_\text{e}$.  Under the Born--Oppenheimer approximation, the relatively heavy atomic nuclei are treated as classical point charges, leaving the electrons to be modeled as quantum particles \cite{born1927quantum}.

Given a set of orthonormal spin-orbitals $\braces{\phi_j}_{j=1}^n$, second quantization and the Jordan--Wigner transformation together map $H_\text{e}$ to a Hamiltonian $H$ in an $n$-qubit state space \cite{mcardle2020quantum}, whose computational basis corresponds with occupancy configurations, known as \textit{Slater determinants}, of the $n$ spin-orbitals. The smallest eigenvalue of $H$, also called the ground state energy, approximates the energy of the true molecular ground state. We consider $H$ as a real linear combination of Pauli strings:
\begin{equation}
\label{eqn:pauli_ham}
    H = \sum_j \omega_j \paren{\sigma^j_1\otimes\dots\otimes\sigma^j_n},\;\;\;\sigma^j_k\in\braces{I,X,Y,Z}
\end{equation}
where the set $\braces{I,X,Y,Z}$ denotes the four Pauli matrices. 
% \[X = \begin{bmatrix}
% 0 & 1\\
% 1 & 0\\
% \end{bmatrix},\;\;\;Y = \begin{bmatrix}
% 0 & -i\\
% i & 0\\
% \end{bmatrix},\;\;\;Z = \begin{bmatrix}
% 1 & 0\\
% 0 & -1\\
% \end{bmatrix},\;\;\;I = \begin{bmatrix}
% 1 & 0\\
% 0 & 1\\
% \end{bmatrix}.\] 
These strings form an orthonormal basis for the $4^n$-dimensional space of $n$-qubit Hamiltonians; the construction of $H$ ensures that the number of terms in \cref{eqn:pauli_ham} is $O(n^4)$ \cite{mcardle2020quantum}. To find the ground state, we consider the expectation value of $H$ with respect to $\ket{\psi_\theta}$, an $n$-qubit variational ansatz:
\begin{equation}
    L(\theta) = \bra{\psi_\theta}H\ket{\psi_\theta}
    \label{eqn:cost_fn}
\end{equation} 
By the Rayleigh--Ritz principle, the only minimum for \cref{eqn:cost_fn} is the ground state energy of $H$, in which case $\ket{\psi_\theta}$ represents an associated eigenvector. The goal is then to variationally optimize $\theta$ to find this ground state. %In quantum computing, $\theta$ is a set of gate parameters that physically prepare the ansatz, and this optimization is known as the variational quantum eigensolver (VQE). Explicitly modeling this problem classically is intractable at scale, but a promising alternative is to model the ansatz implicitly with a neural network, also known as neural quantum states.

\subsection{Autoregressive Neural Quantum States}
\label{sec:background_nqs}

\begin{figure}[t]
    \centering
    \includegraphics[width=0.5\linewidth]{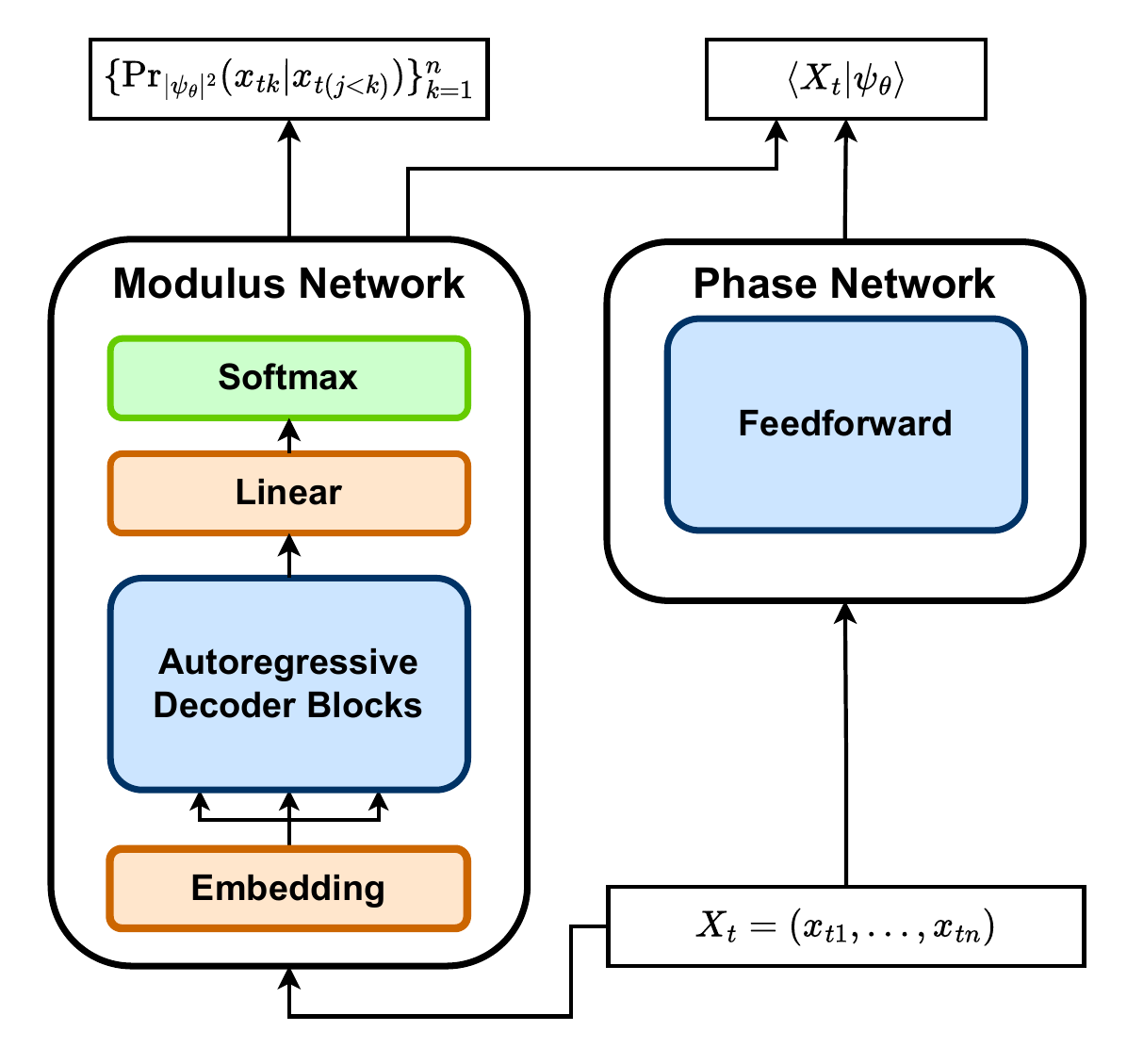}
    \caption{A general autoregressive NQS ansatz $\ket{\psi_\theta}$. The model takes a qubit configuration $x$ and returns either the ansatz entry $\ip{x|\psi_\theta}$ or conditional probabilities $\operatorname{Prob}(x_j \;|\; x_{k<j})$. The modulus network is an autoregressive model that directly contributes to both outputs, while the phase network is typically an MLP only contributing to the ansatz entry.}
    \label{fig:autoregressive_ansatz}
\end{figure}

% \begin{figure}
%     \centering
%     \includegraphics[width=0.3\linewidth]{figures/LLM-NNQS.pdf}
%     \caption{A general autoregressive NQS ansatz $\ket{\psi_\theta}$. The model takes a qubit configuration $x$ and returns either the ansatz entry $\ip{x|\psi_\theta}$ or conditional probabilities $\operatorname{Prob}(x_j \;|\; x_{k<j})$. The modulus network is an autoregressive model that directly contributes to both outputs, while the phase network is typically an MLP only contributing to the ansatz entry.}
%     \label{fig:autoregressive_ansatz}
% \end{figure}

Neural quantum states \cite{carleo2017solving} is a classical deep learning architecture for solving variational many-body problems using a neural network ansatz. Over the past few years, serious interest has manifested toward addressing electronic ground state problems with NQS \cite{Choo_2020}, particularly using autoregressive ansatze \cite{barrett2022autoregressive, knitter2024retnet_nqs, wu2023nnqstransformer, zhao2023nnqsmade}; we exclusively consider the autoregressive formulation in this paper. An $n$-qubit NQS ansatz is a neural network that maps bit strings of length $n$ to complex numbers. For the purposes of training, we consider these outputs as the complex logarithm of the ansatz:
\begin{equation}
    \log\ip{x|\psi_\theta} = \log\psi_\theta(x):\braces{0,1}^n\rightarrow\mathbb{C}
    \label{eqn:nqs_ansatz}
\end{equation}
Autoregressive ansatze encode these values in two networks: a phase net $\phi_\theta$ returning a single number, and a modulus net returning an autoregressive sequence of nested conditional probabilities: \[\log p_\theta^j(x_1,\dots,x_j) = \log\paren{\operatorname{Pr}\paren{x_j = 1|x_{k<j};\theta}}\] The individual network outputs are collected into the ansatz output
\begin{equation}
    \log\ip{x|\psi_\theta} = \frac 1 2\sum_{j=1}^n p_\theta^j(x_{k\leq j})+ i\phi_\theta(x)
    \label{eqn:conditional_prob}
\end{equation}
This specific construction ensures the ansatz always encodes a normalized quantum state, which is significant because every state induces a probability distribution on the set of qubit spin configurations, \[\operatorname{Pr}\paren{x;\theta}=\abs{\ip{x|\psi_\theta}}^2\]

and we may rewrite the expectation \cref{eqn:cost_fn} as a stochastic loss function with respect to this distribution:
\begin{equation}
    L(\theta) =\mathbb{E}_{x\sim\abs{\psi_\theta}^2}\bracket{l_\theta(x)},\;\;\;l_\theta(x)=\frac{\bra{x}H\ket{\psi_\theta}}{\ip{x|\psi_\theta}}
    \label{eqn:nqs_loss}
\end{equation}
The value $l_\theta(x)$ in \cref{eqn:nqs_loss} is called the \textit{local energy} of $H$ at $x$. The autoregressive ansatz construction makes efficient sampling from $\abs{\psi_\theta}^2$ possible, through use of the conditional probabilities generated by the modulus network to sample spins qubit-by-qubit. For each sample $x$, the local energy is constructed by  multiplying all nonzero entries in row $x$ of $H$ with the corresponding entries of $\ket{\psi_\theta}$, then adding the results. In practice, the set of unique samples is stored and a list of frequencies monitors how often each is sampled with replacement \cite{zhao2023nnqsmade}. Explicit constraints are built into the network to ensure the state distribution only samples from physically plausible Slater determinants: the full Hilbert space considers all orbital occupancy configurations, but most do not correspond with the correct number of electrons in the molecule. As in \cref{eqn:nqs_loss}, we can rewrite the gradient of the expectation as a stochastic estimator over the state distribution:
\begin{equation}
    \label{eqn:nqs_gradient}
    \nabla_\theta L(\theta) =  2\operatorname{Re}\mathbb{E}_{x\sim\abs{\psi_\theta}^2}\bracket{\paren{l_\theta(x) - b}\nabla_\theta\log\ip{\psi_\theta|x}}
\end{equation}
The baseline $b$ is arbitrary but typically chosen as $b=L(\theta)$ for purposes of variance reduction.

\subsection{Types of Autoregressive NQS Ansatze}

\begin{table*}[ht!]
\centering
\scriptsize
\begin{tabular}{lll}
\toprule
Model & Sequential Forward FLOP Estimate & Full Ansatz Training FLOP Estimate\\
\midrule
MADE & $3N$ & $BT\big(\paren{1.5n + 3M + 10 - 3\log B}N_\text{mod} +2(M+3)N_\text{ph}\big)$\\[0.5em]
RetNet & $n_\text{seq}\paren{2N + 4n_\text{b}n_\text{seq}d_\text{attn}}$ (Parallel)& $BT\big(\paren{(M+4)n - 2\log B}N_\text{mod} + 2(M+3)N_\text{ph}$\\[0.3em]
&$n_\text{seq}\paren{2N + 5n_\text{b}(d_\text{attn})^2}$ (Recurrent) & $ + n_\text{b}d_\text{m} \times (2.5d_\text{m}\paren{(M+1)n - 2\log B} + 3n^2)\big)$\\[0.5em]
Transformer &$n_\text{seq}\paren{2N + 4n_\text{b}n_\text{seq}d_\text{attn}}$ & $BT\big(\paren{(M+4)n - \tfrac 4 3 \log B}N_\text{mod} + 2(M+3)N_\text{ph}$\\[0.3em]
& & $+ n_\text{b}d_\text{m}\paren{(M+3.5)n^2 - \tfrac 8 9 \log B - \tfrac 2 3 (\log B)^2}\big)$\\
\bottomrule
\end{tabular}
\vspace{0.5em}
\caption{Estimates of the total FLOP counts required both to perform one forward pass of MADE, RetNet, and transformer on an input sequence, based on parameter count $N =N_\text{mod} + N_\text{phase}$, sequence length $n_\text{seq}$, number of decoder blocks $n_\text{b}$, and the attention/retention dimension $d_\text{attn}$; and to train a corresponding NQS ansatz, based on qubit number $n$, training steps $T$, average number $B$ of unique samples processed, and unique bit flip terms $M$ in the problem Hamiltonian. RetNet and Transformer estimates require the number $n_\text{b}$ of blocks and the internal dimension $d_\text{m}$.}
\label{tab:flop_count_formulas}
\end{table*}

% \begin{table}[ht!]
%     \centering
%     \scriptsize
%     \begin{tabular}{cc}
% \toprule
%        Autoregressive Model  & Sequential Forward FLOP Estimate\\
%        \midrule
%         MADE & $3N$\vspace{0.2em}\\
%         Transformer & $n_\text{seq}\paren{2N + 4n_\text{b}n_\text{seq}d_\text{attn}}$\vspace{0.2em}\\
%         RetNet (Parallel) & $n_\text{seq}\paren{2N + 4n_\text{b}n_\text{seq}d_\text{attn}}$\vspace{0.2em}\\
%         RetNet (Recurrent) & $n_\text{seq}\paren{2N + 5n_\text{b}(d_\text{attn})^2}$\\
% \bottomrule
%     \end{tabular}
%     \vspace{0.5em}
%     \caption{Leading term estimations of the total FLOP count required for the MADE, RetNet, and Transformer to perform one forward pass on a sequence of input tokens, based on parameter count $N$, sequence length $n_\text{seq}$, number of decoder blocks $n_\text{b}$, and the attention/retention dimension $d_\text{attn}$.}
%     \label{tab:autoregressive_flop_counts}
% \end{table}

Figure \ref{fig:autoregressive_ansatz} depicts a general autoregressive NQS ansatz, highlighting both modulus and phase networks. The three ansatze examined in this work all follow this form, using a simple feedforward network to model the phase, which is joined by a sequence of autoregressive decoder blocks modeling the conditional probabilities. They differ only in the type of decoder used: in this paper, we consider the MADE \cite{zhao2023nnqsmade}, RetNet \cite{knitter2024retnet_nqs}, and transformer \cite{wu2023nnqstransformer} architectures. Reserving a more detailed description of them for the appendix, we emphasize here that each model enforces the autoregressive constraint on its outputs in a distinct way. Fundamentally, MADE \cite{germain2015made} follows a basic feedforward architecture; a fixed binary mask is applied to the model weights in each forward pass, severing connections between neurons so as to ensure the autoregressive property. This makes MADE easy to sample from and train in a parallelizable way, at the expense of making it a fundamentally less expressive architecture relative to model size.

Transformers \cite{vaswani2017attention} calculate all pairwise dot product similarities within each input sequence. Attention can be made autoregressive through input masking, and is highly parallelizable, though at the expense of introducing a quadratic computational cost with respect to input sequence length. RetNets \cite{sun2023retnet} are recurrent architectures, which are innately autoregressive and perform forward passes in linear time. RetNets compress into a mathematically equivalent parallel form analogous to the transformer; utilizing these two forms during inference and training respectively allows RetNet to utilize both the time complexity advantage of recurrent networks and the greater ease of training of transformers. We describe in \cref{tab:flop_count_formulas} leading term estimates for the FLOP count required by each of these models to perform a single forward pass on a single input sequence, from which it is possible to extrapolate FLOP count estimations for an entire NQS training run. We provide detailed derivations of these estimates in Appendix \ref{sec:appendix_flop_count}.

\section{Methodology}
\label{sec:background_scaling_laws}
We introduce a general framework for FLOP count estimation to quantify the computational cost of NQS. Drawing on the Chinchilla laws \cite{chinchilla}, we present a parametric scaling law for NQS and describe the specific experimental setup of our analysis.

%We introduce a novel general FLOP count estimation framework for approximating the compute cost of NQS. We then introduce a parametric scaling law for NQS inspired by the Chinchilla laws \cite{chinchilla}, and then describe the specific experimental setup of our analysis.

\subsection{FLOP Estimates for NQS}
\label{sec:nqs_flop_count} 

% \begin{table}[ht!]
%     \centering
%     \scriptsize
%     \begin{tabular}{cc}    
% \toprule
%        Ansatz  & FLOP Estimate Formula  \\
%        \midrule
%         MADE & $BT\paren{\paren{1.5n + 3M + 10 - 3\log B}N_\text{mod} + 2(M+3)N_\text{ph}}$\vspace{0.2em}\\
%         RetNet & $BT\big(\paren{(M+4)n - 2\log B}N_\text{mod}+2(M+3)N_\text{ph}$\\
%         & $+ n_\text{b}d_\text{m}\paren{2.5d_\text{m}\paren{\paren{M+1}n - 2\log B} + 3n^2}\big)$\vspace{0.2em}\\
%         Transformer & $BT\Big(\paren{\paren{M+4}n - \frac 4 3 \log B}N_\text{mod} + 2\paren{M+3}N_\text{ph}$\\
%    & $+ n_\text{b}d_\text{m}\paren{\paren{M+3.5}n^2 - \frac 8 9 \log B - \frac 2 3 \paren{\log B}^2}\Big)$\\
% \bottomrule
%     \end{tabular}
%     \vspace{0.5em}
%     \caption{Leading term estimations of the total FLOP count required to train each NQS ansatz, based on qubit number $n$, training steps $T$, average number $B$ of unique samples processed, unique bit flip terms $M$ in the problem Hamiltonian, and parameter count $N_\text{mod} + N\text{phase}$. RetNet and Transformer estimates require the number $n_\text{b}$ of decoder blocks and the internal decoder dimension $d_\text{m}$.}
%     \label{tab:flop_count_formulas}
% \end{table}

Typically, supervised learning only requires one forward and one backward pass through the model for each batch sample at each gradient update step. NQS, however, does not have training data, instead using the model to first generate a batch of sample spins from $x\sim\abs{\psi_\theta}^2$, and then calculate local energies. For each sample $x$, this requires not only one forward pass to obtain $\ip{x|\psi_\theta}$, but many more to calculate the numerator $\ip{x|H|\psi_\theta}$. Each term $P_i$ in \cref{eqn:pauli_ham} defines a specific bit flip sequence: applying this flip to $x$ yields the index $x'$ of the term $\ip{x'|\psi_\theta}$ that corresponds with a nonzero entry of $P_i$. Each bit flip adds an additional forward pass to the compute cost. Before training, all unique bit flip sequences are identified between the terms of $H$, eliminating redundant forward passes \cite{wu2023nnqstransformer}. Nonetheless, a significant number of forward passes remains, though none of them contribute to the gradient calculation, yielding some compute cost savings depending on the ansatz.

For an NQS problem, we estimate the training FLOP count based on the number $n = 2n_\text{seq}$ of qubits, the average unique sample batch size $B$ at each iteration, the single-token FLOP count $F_\text{mod}$ of the modulus network, the single-sequence FLOP count $F_\text{ph}$ of the phase network, the number $M$ of unique bit flip patterns within $H$, and the training steps $T$. We estimate the sampling FLOP count as
\begin{equation}
    \operatorname{FLOP}_\text{Sample} \approx F_\text{mod}T\paren{\sum_{m=0}^{\lfloor\log_4 B\rfloor - 1}4^m + B\paren{\frac n 2 - \lfloor\log_4 B\rfloor}}
    \label{eqn:sample_flop}
\end{equation}

For ease of approximation, we assume at each step the sampling process starts from the four configurations of a single orbital and branches into $B$ unique samples in approximately $\log_4(B)$ steps. The remaining steps are needed to identify the full spin configuration for each sample. This assumption is an overestimate, since the sampling process is less costly if it does not branch this quickly. Then, the FLOP count to construct and backpropagate through \cref{eqn:nqs_loss} is approximately

\begin{equation}
    \operatorname{FLOP}_\text{Loss} \approx BT\paren{\paren{M+3}\paren{\frac n 2F_\text{mod} + F_\text{ph}}}
    \label{eqn:loss_flop}
\end{equation}

For each sample at each time step, \cref{eqn:loss_flop} covers $M$ forward passes for the local energy numerator, one for the denominator, and then backpropagation---approximately equivalent to two forward passes. We ignore the FLOP count for assembling the model outputs into loss values as they contribute no leading terms to the overall count.  We also note that $F_\text{mod}$ is implicitly context-dependent even within \cref{eqn:sample_flop,eqn:loss_flop}; for each ansatz, $F_\text{mod}$ may vary based on the corresponding phase of training.

From derivations provided in \cref{sec:appendix_flop_count}, we arrive at the ansatz-specific FLOP count estimates for MADE, RetNet, and transformer shown in \cref{tab:flop_count_formulas}. For the latter two ansatze, we set $d_\text{attn}=d_\text{m}$. In all cases, we consider separately the number of parameters $N = N_{\text{mod}} + N_\text{ph}$ for the modulus and phase networks, respectively, ignoring all non-leading terms. KV caching can reduce some of the transformer compute cost, for a considerable increase in memory. Accomplishing this cost savings would also require cache sharing between separate forward passes to an atypical extent. Though this paper performs FLOP-constrained scaling law analysis using rougher estimates given in \cref{tab:revised_flop}, which have been simplified for ease of analysis, we introduce these three estimate formulas here as general resources for assessing the compute cost of the MADE, transformer, and RetNet ansatze.

\subsection{Parametric Scaling Law for NQS}
\label{subsec:parameteric_curves}

We now derive NQS scaling laws to identify predictive relationships between accuracy/performance and several key attributes of NQS: model size, compute budget, problem size, and the number of unique samples used for training. In practice, the search space grows too quickly with problem size to allow for sampling from all feasible spins under a fixed compute budget, so we truncate the number of unique samples generated at each sampling run. Since these samples contribute to the loss function analogously to training data, the total number obtained in training serves as a proxy for the number of tokens used in traditional scaling law analysis.

Appendix \ref{sec:appendix_chinchilla_curve} gives a brief overview of the Chinchilla scaling law \cite{chinchilla} establishing basic power law relationships between the final training loss $L$, model size $N$, and number $D$ of tokens used. To adapt this curve to NQS, we must consider that any scaling analysis predicting the general behavior of an ansatz across a variety electronic ground state calculations must account for performance differences caused by training on different molecules. Increasing problem size, while keeping all other hyperparameters equal, is expected to both increase computational cost and decrease accuracy for any ansatz. To help the scaling law account for variation in problem size, we propose incorporating, in addition to the number $N$ of model parameters (per thousand), a new term $SF$ equaling the average fraction of the total search space that is sampled throughout training. For a molecular system with $n$ spin-orbitals, $N_E$ electrons, and a singlet ground state---where the electrons are equally divided among the up and down spins---the search space size is $\genfrac{(}{)}{0pt}{0}{0.5n}{0.5N_E}^2.$

This calculation is analogous for molecules with nonzero spin, though all molecules used for analysis in \cref{sec:experimental_setup} have either singlet or triplet ground states. For a given NQS run, we then obtain $SF$ by dividing the average number $B$ of unique samples generated at each iteration by the search space size. Since $SF$ incorporates information about the number of unique samples seen at each optimization step, we multiply it by the number $T$ of optimization steps to give \[D' = T\times SF\] a size-controlled analog to the variable $D$ from \cite{chinchilla}.

Since different molecules have different energies, we modify the loss $L(N,D')$: absolute error between the NQS result and the ground truth, obtainable for moderate problem sizes via full configuration interaction (FCI), is a suitable measure since an error tolerance within $0.00159$ Hartree is a widely accepted standard of chemical accuracy. Unfortunately, absolute error is not viable at scale since FCI calculations quickly become intractable. As such, we also consider the V-Score \cite{wu2024variational}, a recent performance metric for assessing variational Hamiltonian eigensolvers. For an $n$-qubit Hamiltonian $H$ and ansatz $\ket{\psi_\theta}$, the V-Score is given by
\begin{equation}
    \operatorname{V-Score} = n\frac{\ip{\psi_\theta |H^2| \psi_\theta}-\ip{\psi_\theta |H| \psi_\theta}^2}{\big({\ip{\psi_\theta |H| \psi_\theta} - \omega_0}\big)^2}
    \label{eqn:v_score}
\end{equation}
where $\omega_0$ is the weight value from \cref{eqn:pauli_ham} corresponding with identity operator. Both \cref{eqn:cost_fn} and its variance---the numerator in \cref{eqn:v_score}---may be estimated by the sample mean and variance of \cref{eqn:nqs_loss}. V-Score was developed as a proxy measure for relative error, and it is both dimensionless and invariant with respect to global energy shifts. These properties make V-Score a natural candidate for scaling law analysis. Altogether, we consider two modifications of the Chinchilla curve for assessing the performance of NQS models for electronic ground state calculations: 
\begin{equation}
    \operatorname{Error_{Abs}}(N,D') = A_0 + \frac{A_1}{N^{\alpha_1}} + \frac{A_2}{D'^{\alpha_2}}, \ \ \operatorname{V-Score}(N,D') = A_0 + \frac{A_1}{N^{\alpha_1}} + \frac{A_2}{D'^{\alpha_2}}
    \label{eqn:vscore_curve}
\end{equation}

\subsection{Experimental Setup}
\label{sec:experimental_setup}

%\textcolor{red}{CAN WE EXPRESS SOME OF THE LARGE VALUES BELOW VIA SCIENTIFIC NOTATION?}
\begin{table}[ht!]
    \centering
    % \small
    \scriptsize
    \begin{tabular}{ccccccc}
        \toprule
        Molecule & $n$ & $N_E$ & Determinants  & Paulis & $M$ & FCI \\
        \midrule
        H\textsubscript{2}O & 14 & 10 & 441 & 1,390 & 190 & $-75.0155$ \\
        N\textsubscript{2} & 20 & 14 & 14,400 & 2,591 & 446 & $-107.6602$ \\
        O\textsubscript{2} & 20 & 16 & 2,025 & 2,879 & 530 & $-147.7502$ \\
        H\textsubscript{2}S & 22 & 18 & 3,025 & 9,558 & 1,524 & $-394.3546$ \\
        PH\textsubscript{3} & 24 & 18 & 48,400 & 24,369 & 4,333 & $-338.6984$ \\
        LiCl & 28 & 20 & 1,002,001 & 24,255 & 4,508 & $-460.8496$ \\
        Li\textsubscript{2}O & 30 & 14 & 41,409,225 & 20,558 & 3,810 & $-87.8927$ \\
        \bottomrule
    \end{tabular}
    \vspace{0.5em}
    \caption{Molecules whose second quantized (in the the STO-3G basis set) Hamiltonians were used for analysis. We list the corresponding number of spin-orbitals, electrons, Slater determinants in the search space, Pauli basis terms, and unique bit flip patterns among the Pauli terms. FCI ground truth values from \cite{wu2023nnqstransformer} are also shown.}
    \label{tab:scaling_data_molecules}
\end{table}

We apply each ansatz to each of the molecules in \cref{tab:scaling_data_molecules}, performing a coarse ablation varying model size---both modulus and phase networks---optimization steps, and the maximum number of unique samples generated per step. We perform all training with Adam, annealing the learning rate along a cosine schedule from $2.5\times 10^{-3}$ to $5\times 10^{-8}$, with a linear warmup over the first $4\%$ of steps. Following the same practice from \cite{knitter2024retnet_nqs}, we exponentially increase the number of non-unique samples at each step from $10^4$ to $10^{12}$, and for the first 90\% of training, we only recompute the loss every 10 steps. We also incorporate variational neural annealing (VNA) \cite{hibat2021variational} to improve accuracy. We vary the number of training steps between 12,500 and 50,000, the model dimensions of RetNet and transformer between 8 and 64---the internal feedforward dimension always equals 4x this dimension---and the hidden state dimensions of the MADE and phase networks between 8 and 128. For both RetNet and transformer, we utilized 1 or 2 decoder blocks. The maximum number of unique samples generated at each step was varied between 1,000 and 16,000.

We follow \cite{knitter2024retnet_nqs} in choosing the final loss value of each training run. We fit Eq. \cref{eqn:vscore_curve} to our data as in \cite{chinchilla}, replacing zero absolute error values with $10^{-5}$ to avoid issues with the log-loss and using Adam alongside a cosine-annealed learning rate for 50,000 steps, with initial parameters taken from ranges given for the original Chinchilla law \cite{besiroglu2024chinchilla, chinchilla}. 
% Several curve fitting trials are performed with the best ones kept in lieu of an exhaustive search.

\section{Results}
\label{sec:result}

\begin{figure}
    \centering
    \includegraphics[width=0.95\linewidth]{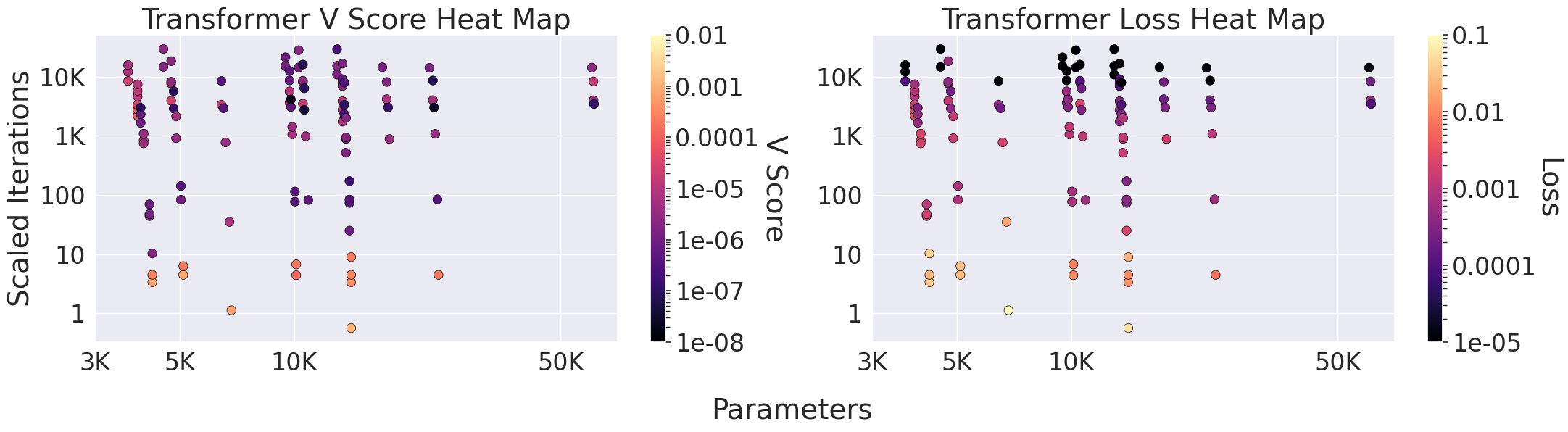}
    \caption{Heat maps showing the final V-score and loss values for each data point collected for the transformer ansatz, as a function of both model parameters $N$ and scaled iterations $D'$. Since the variance and loss values comprising the V-score are stochastic estimates, we depict $10^{-8}$ as the lower limit of V-score accuracy, while absolute loss values are truncated at $10^{-5}$ to reflect desired thresholds of chemical accuracy.}
    \label{fig:transformer_heatmaps}
\end{figure}

We collected 335 datapoints applying the three ansatze to the problems in \cref{tab:scaling_data_molecules}: 120 for MADE, 109 for RetNet, and 106 for transformer. We observe in \cref{fig:transformer_heatmaps} the distribution of data points collected for transformer, to which those of MADE and RetNet are qualitatively similar and given in \cref{sec:appendix_heatmaps}. These heat maps display the V-scores and absolute loss values as functions of the variables $N$ and $D'$ as defined in \cref{subsec:parameteric_curves}: the figures qualitatively indicate a correlation between $N$, $D'$, and both dependent variables. Similar trends are observable within the data sets for both the tranformer and RetNet ansatze. Reserving some discussion of these other models for the appendix, we emphasize how similar the RetNet behavior in \cref{fig:retnet_heatmaps} is to that of the transformer shown here. In general, the data point distributions of the two networks are directly comparable, since RetNet and transformer ansatze of equivalent dimension have nearly identical numbers of parameters; likewise, both models train similarly under equivalent setups \cite{knitter2024retnet_nqs}.

In both cases, we observe a pronounced stratification of the data point distribution with respect to $N$. A similar stratification is visible in \cref{fig:made_heatmaps}, but that data set succeeds in covering a wider range of parameter sizes. This difference is due to the fact that the architecture of RetNets and transformers is less explicitly dependent on problem size than that of MADE. Furthermore, in testing, we maintained a fixed ratio of 8 hidden dimensions per attention/retention head inside each transformer and RetNet.

\begin{table}[ht!]
    \centering
    \scriptsize
    \begin{tabular}{ccccccc}
    \toprule
        V-score & $A_0$ & $A_1$ & $A_2$ & $\alpha_1$& $\alpha_2$ & Log-$R^2$\\
        \midrule
        MADE & $9.37\times 10^{-11}$ & $2.58 \times 10^{-5}$ & $5.53 \times 10^{-2}$ &  $1.459$ & $2.828$ & $0.68$ \\
        RetNet & $6.71\times 10^{-7}$ & $0.626$ & $4.39 \times 10^{-3}$ & $9.370$& $2.451$ & $0.40$\\
        Transformer & $1.34\times10^{-6}$ & $0.197$ & $7.25 \times 10^{-3}$ & $7.952$& $2.371$ & $0.53$\\
        \midrule
        Absolute Error & $A_0$ & $A_1$ & $A_2$ & $\alpha_1$& $\alpha_2$ & Log-$R^2$\\
        \midrule
        MADE & $4.12\times 10^{-9}$ & $0.033$ & $0.106$ & $2.224$ & $0.757$ & $0.65$ \\
        RetNet & $1.13\times 10^{-4}$ & $5.26 \times 10^{-4}$ & $0.111$ & $0.452$& $1.179$ & $0.49$\\
        Transformer & $2.83\times10^{-9}$ & $0.720$ & $0.039$ & $5.274$& $0.637$& $0.62$\\
         \bottomrule
    \end{tabular}
    \vspace{0.5em}
    \caption{Parametric curve parameters for MADE, RetNet, and transformer NQS ansatze, for both V-score and Loss data. $R^2$ are given on a log--log scale between model outputs and true loss values.}
    \label{tab:nqs_scaling_curve_results}
\end{table}

Following the training procedure from \cref{sec:experimental_setup}, we have identified parametric curve coefficients, shown in \cref{tab:nqs_scaling_curve_results} for \cref{eqn:vscore_curve} for all three ansatze. These coefficients indicate that model size makes a greater impact, as represented by $A_1$ and $\alpha_1$, on V-score for the transformer and RetNet models, although this preference is not as strong for the transformer. In contrast, model size appears to make a nearly negligible impact for MADE, and here we should rather focus on increasing training time as a means of improving performance. The transformer ansatz exhibits scaling behavior that is more similar to RetNet, preferring an increase in model size to an increase in training time.

Looking now to the coefficients for the absolute error values, we immediately observe that the baseline coefficient $A_0$ is smaller than the predetermined error tolerance for both MADE and transformer. This observation alone is not alarming, as all datasets contain several problem instances where the model achieved the exact tolerance. For the RetNet, a higher baseline coefficient may indicate the model is fundamentally less accurate than the other two, but such a distinction is not practically meaningful since the baseline still lies within chemical accuracy. Furthermore, these curves indicate that model size impacts final accuracy less for the RetNet, so one should focus on increasing training time; in contrast, it appears more advantageous to increase model size for the transformer.

\subsection{FLOP Analysis}
\begin{figure}
    \centering
    \includegraphics[width=0.95\linewidth]{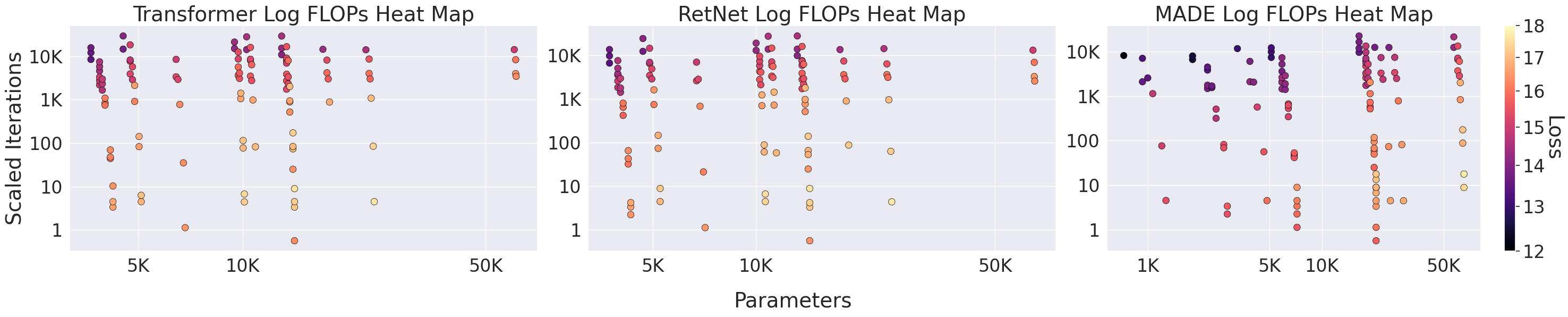}
    \caption{Heat maps showing the logarithm of the FLOP count estimates for all data points collected with each of the three ansatze. In all three cases, FLOP count does not decrease monotonically with respect to scaled iterations, indicating the latter is not merely a proxy for problem size.}
    \label{fig:log_flop_heatmaps}
\end{figure}

In \cref{fig:log_flop_heatmaps}, we show FLOP estimates, on a logarithmic scale, for each of the entries in our three datasets. These estimates come from the equations outlined in \cref{sec:nqs_flop_count}, and provide a coarse, yet still meaningful, understanding of the compute time needed to produce these results. Qualitatively, we observe for all three ansatze that when controlling for scaled training time $D'$, an increase in model size $N$ correlates with an increase in compute time, which makes intuitive sense.

Conversely, $D'$ scales inversely with compute requirements; this observation is also intuitive, since larger problem sizes and search spaces should obviously correlate with larger compute costs: for a fixed training time, a larger problem size will generally incur a greater compute cost and lower $D'$. As a result, one may consider the argument that $D'$ only serves to predict the V-score and absolute error values in our data to the extent that smaller problems are easier to learn than larger ones. As a potential rebuttal, and an indication that $D'$ may reveal something more fundamental, we observe in \cref{fig:log_flop_heatmaps} for all ansatze that when keeping $N$ fixed, an increase in $D'$ appears to first correspond with an increase in FLOPs before the general decrease: a similar reversal does not appear in either the V-score or absolute error data. This phenomenon indicates that the correlation between $D'$ and FLOPs is likely weaker than that between $D'$ and our two loss metrics, lending credence to the hypothesis that $D'$ is truly a size-agnostic measure of training time.

\subsection{Compute Constrained Scaling Analysis}
\begin{table}[ht!]
    \centering
    \begin{tabular}{cc}
    \toprule
       Ansatz  & Simplified Training FLOP Estimate\\
    \midrule
    MADE & $3MSD'N$\\
    RetNet & $\frac 1 {13}\paren{15.5Mn + \frac{3n^2}{d_\text{m}}}SD'N$\\
        Transformer & $M\paren{n + \frac{n^2}{12d_\text{m}}}SD'N$\\
    \bottomrule
    \end{tabular}
    \vspace{0.5em}
    \caption{Simplified FLOP estimates for constrained scaling law optimization. These are derived from the estimates given in \cref{tab:flop_count_formulas} and are intended to prioritize ease of computation at the expense of a decrease in accuracy. Here $S$ refers to the search space size of the NQS problem.}
    \label{tab:revised_flop}
\end{table}
The FLOP estimates in \cref{tab:flop_count_formulas}, while already ignoring many non-leading contributions, are meant to reflect the primary contributions of all relevant factors of an NQS training run on compute cost. These complicated formulas contrast with the simple $6ND$ cost formula assumed by \cite{kaplan2020scaling} for language modeling, which allows for a simple constrained optimization comparing performance relative to compute cost. To perform a similar analysis here, we introduce further simplified compute estimates in \cref{tab:revised_flop}, with detailed discussion of their derivation in \cref{sec:flop_constrained_derivation}. Future analyses that consider $N_\text{mod}$ and $N_\text{ph}$ as separate variables may benefit from the more accurate estimates of \cref{tab:flop_count_formulas}.

\begin{table}[ht!]
    \centering
    \begin{tabular}{cccc}
    \toprule
       $D'=$  & MADE & RetNet & Transformer \\
    \midrule
        V-score & $0.053 \times N^{0.516}$ & $13.076 \times N^{3.823}$ & $6.707 \times N^{3.354}$ \\
        Absolute Error & $0.889 \times N^{2.938}$ & $0.005 \times N^{0.383}$ & $2663.825 \times N^{8.279}$ \\
    \bottomrule
    \end{tabular}
    \vspace{0.5em}
    \caption{The empirical coefficients in \cref{tab:nqs_scaling_curve_results} and estimates in \cref{tab:revised_flop} together define curves tracking the growth of $D'$, with respect to $N$, along the efficient frontier determined by compute budget $C$.}
    \label{tab:efficient_frontier}
\end{table}

Derivations made in \cref{sec:constrained_optimization} indicate that, for compute budget estimates of the form $C=kD'N$ for some problem-defined $k$, such as those given in \cref{tab:revised_flop}, the constrained optimization of \cref{eqn:vscore_curve} is entirely analogous to the process carried out in \cite{chinchilla}. Consequently, we give in \cref{tab:efficient_frontier} the optimal growth rate of $D'$ as a function of $N$, for each loss metric and ansatz considered. These growth rates are highly metric and ansatz dependent, and more importantly, do not reflect the approximately linear curves identified in the Chinchilla scaling laws.

\section{Conclusion}
\label{sec:conclusion}

% Neural quantum states represents a promising avenue for calculating electronic ground states accurately and efficiently. Much progress has been made to demonstrate the fundamental utility of NQS in this space, but significant work remains to better understand the exact scaling behaviors of specific NQS architectures.
Our work applies Chinchilla-style LLM scaling laws, incorporating a new training time variable dependent on problem size, to MADE, transformer, and RetNet NQS ansatze in order to assess their performance on a selection of quantum chemistry problems of varying size. Using an ablation dataset, we fit this curve to predict both absolute error and V-score loss metrics for each ansatz. Together with a novel framework for estimating the compute cost of NQS, these results demonstrate how the performance of NQS at scale may be analyzed similarly to LLMs.

This line of research invites a more comprehensive ablation study, with greater variety of architectures, problem sizes and training setups. For instance, the model size $N$ in \cref{eqn:vscore_curve} comprises both modulus and phase networks, but it may be illuminating to isolate the phase network's role. Varying the total number of non-unique Monte Carlo samples may also impact accuracy.
% The primary obstacle here would be the time and resources required to produce data. A potential solution may lie in analyzing variations of NQS intended to reduce computational time. Current FLOP count estimates are based on exact local energy calculations.
A technique introduced in \cite{wu2023nnqstransformer} efficiently approximates local energy calculations by neglecting Hamiltonian terms associated with unsampled states, and more recent work \cite{malyshev2024neuralquantumstatespeaked} demonstrates the extent of its potential. Analyzing the scaling properties of NQS under this new variation would help provide insight into the extent of its benefits, especially when considering NQS as a problem-agnostic method of dequantization \cite{knitter2022vnls}.

\section*{Acknowledgments}
The authors would like to thank James Stokes for his feedback throughout the course of this work.

\bibliographystyle{plain}
\bibliography{references}

\begin{thebibliography}{10}

\bibitem{barrett2022autoregressive}
Thomas~D Barrett, Aleksei Malyshev, and AI~Lvovsky.
\newblock Autoregressive neural-network wavefunctions for ab initio quantum chemistry.
\newblock {\em Nature Machine Intelligence}, 4(4):351--358, 2022.

\bibitem{bartlett2007coupled}
Rodney~J Bartlett and Monika Musia{\l}.
\newblock Coupled-cluster theory in quantum chemistry.
\newblock {\em Reviews of Modern Physics}, 79(1):291--352, 2007.

\bibitem{bennewitz2022neural}
Elizabeth Bennewitz, Florian Hopfmueller, Bohdan Kulchytskyy, Juan Carrasquilla, and Pooya Ronagh.
\newblock Neural error mitigation of near-term quantum simulations.
\newblock {\em Nature Machine Intelligence}, 4(7):618--624, Jul 2022.

\bibitem{besiroglu2024chinchilla}
Tamay Besiroglu, Ege Erdil, Matthew Barnett, and Josh You.
\newblock Chinchilla scaling: A replication attempt.
\newblock {\em arXiv preprint arXiv: 2404.10102}, 2024.

\bibitem{born1927quantum}
Max Born and Robert Oppenheimer.
\newblock Zur quantentheorie der molekeln.
\newblock {\em Annalen der Physik}, 389(20):457--484, 1927.

\bibitem{carleo2017solving}
Giuseppe Carleo and Matthias Troyer.
\newblock Solving the quantum many-body problem with artificial neural networks.
\newblock {\em Science}, 355(6325):602--606, 2017.

\bibitem{Choo_2020}
Kenny Choo, Antonio Mezzacapo, and Giuseppe Carleo.
\newblock Fermionic neural-network states for ab-initio electronic structure.
\newblock {\em Nature Communications}, 11(1), May 2020.

\bibitem{scaling_collapse}
Yunzhen Feng, Elvis Dohmatob, Pu~Yang, Francois Charton, and Julia Kempe.
\newblock A tale of tails: Model collapse as a change of scaling laws.
\newblock In {\em ICLR 2024 Workshop on Navigating and Addressing Data Problems for Foundation Models}, 2024.

\bibitem{germain2015made}
Mathieu Germain, Karol Gregor, Iain Murray, and Hugo Larochelle.
\newblock Made: Masked autoencoder for distribution estimation.
\newblock In {\em International conference on machine learning}, pages 881--889. PMLR, 2015.

\bibitem{scaling_transfer}
Danny Hernandez, Jared Kaplan, Tom Henighan, and Sam McCandlish.
\newblock Scaling laws for transfer.
\newblock {\em arXiv journal arXiv:2102.01293}, 2021.

\bibitem{hibat2021variational}
Mohamed Hibat-Allah, Estelle~M Inack, Roeland Wiersema, Roger~G Melko, and Juan Carrasquilla.
\newblock Variational neural annealing.
\newblock {\em Nature Machine Intelligence}, 3(11):952--961, 2021.

\bibitem{chinchilla}
Jordan Hoffmann, Sebastian Borgeaud, Arthur Mensch, Elena Buchatskaya, Trevor Cai, Eliza Rutherford, Diego de~Las~Casas, Lisa~Anne Hendricks, Johannes Welbl, Aidan Clark, Tom Hennigan, Eric Noland, Katie Millican, George van~den Driessche, Bogdan Damoc, Aurelia Guy, Simon Osindero, Karen Simonyan, Erich Elsen, Jack~W. Rae, Oriol Vinyals, and Laurent Sifre.
\newblock Training compute-optimal large language models.
\newblock {\em arXiv preprint arXiv:2203.15556}, 2022.

\bibitem{mini_scaling}
Shengding Hu, Yuge Tu, Xu~Han, Chaoqun He, Ganqu Cui, Xiang Long, Zhi Zheng, Yewei Fang, Yuxiang Huang, Weilin Zhao, Xinrong Zhang, Zheng~Leng Thai, Kaihuo Zhang, Chongyi Wang, Yuan Yao, Chenyang Zhao, Jie Zhou, Jie Cai, Zhongwu Zhai, Ning Ding, Chao Jia, Guoyang Zeng, Dahai Li, Zhiyuan Liu, and Maosong Sun.
\newblock Minicpm: Unveiling the potential of small language models with scalable training strategies.
\newblock {\em arXiv preprint arXiv:2404.06395}, 2024.

\bibitem{scaling_downstream}
Berivan Isik, Natalia Ponomareva, Hussein Hazimeh, Dimitris Paparas, Sergei Vassilvitskii, and Sanmi Koyejo.
\newblock Scaling laws for downstream task performance of large language models.
\newblock {\em arXiv preprint arXiv:2402.04177}, 2024.

\bibitem{kaplan2020scaling}
Jared Kaplan, Sam McCandlish, Tom Henighan, Tom~B Brown, Benjamin Chess, Rewon Child, Scott Gray, Alec Radford, Jeffrey Wu, and Dario Amodei.
\newblock Scaling laws for neural language models.
\newblock {\em arXiv preprint arXiv:2001.08361}, 2020.

\bibitem{knitter2022vnls}
Oliver Knitter, James Stokes, and Shravan Veerapaneni.
\newblock Toward neural network simulation of variational quantum algorithms.
\newblock {\em NeurIPS workshop on AI for Science: Progress and Promises}, 2022.

\bibitem{knitter2024retnet_nqs}
Oliver Knitter, Dan Zhao, James Stokes, Martin Ganahl, Stefan Leichenauer, and Shravan Veerapaneni.
\newblock Retentive neural quantum states: Efficient ans{\"a}tze for ab initio quantum chemistry.
\newblock {\em Machine Learning: Science and Technology}, 2025.

\bibitem{lin2022scaling}
Sheng-Hsuan Lin and Frank Pollmann.
\newblock Scaling of neural-network quantum states for time evolution.
\newblock {\em physica status solidi (b)}, 259(5):2100172, 2022.

\bibitem{malyshev2024neuralquantumstatespeaked}
Aleksei Malyshev, Markus Schmitt, and A.~I. Lvovsky.
\newblock Neural quantum states and peaked molecular wave functions: Curse or blessing?, 2024.

\bibitem{mcardle2020quantum}
Sam McArdle, Suguru Endo, Al{\'a}n Aspuru-Guzik, Simon~C Benjamin, and Xiao Yuan.
\newblock Quantum computational chemistry.
\newblock {\em Reviews of Modern Physics}, 92(1):015003, 2020.

\bibitem{scaling_dataconstrain}
Niklas Muennighoff, Alexander~M. Rush, Boaz Barak, Teven~Le Scao, Aleksandra Piktus, Nouamane Tazi, Sampo Pyysalo, Thomas Wolf, and Colin Raffel.
\newblock Scaling data-constrained language models.
\newblock {\em arXiv preprint arXiv:2305.16264}, 2023.

\bibitem{beyond_chinchilla}
Nikhil Sardana, Jacob Portes, Sasha Doubov, and Jonathan Frankle.
\newblock Beyond chinchilla-optimal: Accounting for inference in language model scaling laws.
\newblock In {\em Forty-first International Conference on Machine Learning}, 2024.

\bibitem{schollwock2005density}
Ulrich Schollw{\"o}ck.
\newblock The density-matrix renormalization group.
\newblock {\em Reviews of modern physics}, 77(1):259--315, 2005.

\bibitem{sharir2020deep}
Or~Sharir, Yoav Levine, Noam Wies, Giuseppe Carleo, and Amnon Shashua.
\newblock Deep autoregressive models for the efficient variational simulation of many-body quantum systems.
\newblock {\em Physical review letters}, 124(2):020503, 2020.

\bibitem{sun2023retnet}
Yutao Sun, Li~Dong, Shaohan Huang, Shuming Ma, Yuqing Xia, Jilong Xue, Jianyong Wang, and Furu Wei.
\newblock Retentive network: A successor to transformer for large language models.
\newblock {\em arXiv preprint arXiv:2307.08621}, 2023.

\bibitem{scaling_arch}
Yi~Tay, Mostafa Dehghani, Samira Abnar, Hyung~Won Chung, William Fedus, Jinfeng Rao, Sharan Narang, Vinh~Q. Tran, Dani Yogatama, and Donald Metzler.
\newblock Scaling laws vs model architectures: How does inductive bias influence scaling?
\newblock In {\em The 2023 Conference on Empirical Methods in Natural Language Processing}, 2023.

\bibitem{vaswani2017attention}
Ashish Vaswani, Noam Shazeer, Niki Parmar, Jakob Uszkoreit, Llion Jones, Aidan~N Gomez, \L~ukasz Kaiser, and Illia Polosukhin.
\newblock Attention is all you need.
\newblock In I.~Guyon, U.~Von Luxburg, S.~Bengio, H.~Wallach, R.~Fergus, S.~Vishwanathan, and R.~Garnett, editors, {\em Advances in Neural Information Processing Systems}, volume~30. Curran Associates, Inc., 2017.

\bibitem{wu2024variational}
Dian Wu, Riccardo Rossi, Filippo Vicentini, Nikita Astrakhantsev, Federico Becca, Xiaodong Cao, Juan Carrasquilla, Francesco Ferrari, Antoine Georges, Mohamed Hibat-Allah, Masatoshi Imada, Andreas~M. Läuchli, Guglielmo Mazzola, Antonio Mezzacapo, Andrew Millis, Javier~Robledo Moreno, Titus Neupert, Yusuke Nomura, Jannes Nys, Olivier Parcollet, Rico Pohle, Imelda Romero, Michael Schmid, J.~Maxwell Silvester, Sandro Sorella, Luca~F. Tocchio, Lei Wang, Steven~R. White, Alexander Wietek, Qi~Yang, Yiqi Yang, Shiwei Zhang, and Giuseppe Carleo.
\newblock Variational benchmarks for quantum many-body problems.
\newblock {\em Science}, 386(6719):296--301, 2024.

\bibitem{wu2023nnqstransformer}
Yangjun Wu, Chu Guo, Yi~Fan, Pengyu Zhou, and Honghui Shang.
\newblock Nnqs-transformer: an efficient and scalable neural network quantum states approach for ab initio quantum chemistry.
\newblock In {\em Proceedings of the International Conference for High Performance Computing, Networking, Storage and Analysis}, pages 1--13, 2023.

\bibitem{zhao2023nnqsmade}
Tianchen Zhao, James Stokes, and Shravan Veerapaneni.
\newblock Scalable neural quantum states architecture for quantum chemistry.
\newblock {\em Machine Learning: Science and Technology}, 4(2):025034, jun 2023.

\end{thebibliography}

\appendix
\section{Appendix}

\subsection{The MADE Ansatz}
The first ansatz we analyze \cite{zhao2023nnqsmade} utilizes MADE \cite{germain2015made}, a feedforward network designed as an autoencoder for probability distributions, as the decoder. MADE applies binary masks to its weights, severing key neural connections between the hidden layers to ensure that its outputs obey the autoregressive property. MADE does not require any feature embedding or positional encoding of the input spins, and is the simplest of the three architectures being examined. MADE was not the first autoregressive ansatz applied to electronic ground state calculations, but it serves as a precursor for the more sophisticated architectures discussed.

FLOP estimates for a MADE network are simple: a feedforward network with $N$ parameters requires approximately $2N$ FLOPs to perform a forward pass on a single spin configuration, since most model weights correspond with one multiplication and one addition operation inside each forward pass; biases only correspond with one addition operation, but are heavily outnumbered among the total parameter count. We ignore the contribution of nonlinearities to the compute cost, alongside other non-leading FLOP contributions, throughout this work. These count estimates hold true for a MADE network, but the binary masking of model weights introduces an additional FLOP for each weight, so MADE requires $3N$ FLOPs to perform a forward pass on a single spin configuration.

\subsection{NQS--Transformer}
The transformer is a ubiquitous architecture that underpins many large language models. They process input sequences in parallel and generate autoregressive output, making them well-suited to learn the conditional probabilities of an NQS ansatz \cite{bennewitz2022neural, wu2023nnqstransformer}. Each transformer block comprises an attention block and a feedforward layer, interspersed with residual connections and layer normalizations. Attention forms the the key input processing apparatus, calculating a weighted context vector for each input token based on all pairwise contextual relationships within the input sequence. Tokens are embedded based on value and relative position, then projected to obtain batches of query, key, and value vectors, which are used to construct the context vectors:
\begin{equation}
    \operatorname{Attention}(X) = \operatorname{softmax}\paren{\frac{QK^T}{\sqrt{d_k}}}V, \text{ where } Q = XW_Q,\; K = XW_K,\; V=XW_V
    \label{eqn:self_attention}
\end{equation}

Attention can easily learn correlations between tokens separated by significant relative distance, and a transformer typically utilizes several attention heads simultaneously---known as \textit{multi-head attention}---to learn different types of contextual information \cite{vaswani2017attention}. Appropriate input masking during training and inference makes attention autoregressive. Being so naturally expressive, transformers have been shown to perform well as NQS ansatze \cite{wu2023nnqstransformer}. In this context, the transformer operates on input sequences of spatial orbital occupancies. Each spatial orbital can contain two electrons, so these sequences have length $n/2$ for an $n$-qubit NQS problem.

FLOP estimates \cite{kaplan2020scaling} indicate that for a transformer with $N$ total parameters, $n_\text{b}$ blocks, and attention dimension $d_\text{attn}$, a forward pass across $n_\text{seq}$ tokens requires approximately $2N + 4n_\text{b}n_\text{seq}d_\text{attn}$ operations per token. FLOP estimates made for LLMs typically truncate this value to $2N$, but for NQS, the typically smaller ratio of model-to-problem size, coupled with the need for multiple forward passes to generate local energy values, necessitates considering the full FLOP estimate \cite{knitter2022vnls}. We emphasize here that the presence of $n_\text{seq}$ inside this estimate implies that the transformer requires $O(n_\text{seq}^2)$ time to perform a forward pass an entire sequence.

\subsection{RetNet NQS}

The Retentive Network (RetNet) is a recurrent language model that compresses into an equivalent parallel form \cite{sun2023retnet}, which processes inputs and calculate gradients through backpropagation similarly to transformers. In this format, the RetNet block is essentially analogous to the transformer, except that multi-head attention is replaced by multi-scale retention, which compresses the repeated action of a linear recurrent model onto the embedded inputs:
\begin{equation}
    \operatorname{Retention}(X) = \paren{QK^T\odot D}V, \text{ where } Q = (XW_Q)\odot\Theta,\; K = (XW_K)\odot\overline\Theta,\; V=XW_V.
    \label{eqn:parallel_retention}
\end{equation}

Under retention, embedded tokens are still projected to queries, keys, and values, but the retention head itself encodes relative positional information $\Theta$, along with applying its own weighted autoregressive mask $D$. These matrices are defined by
 \begin{equation}
     \label{eqn:theta_and_D}
     \Theta_j = e^{ij\theta},\;\;\; D_{jk} = \begin{cases}
         \gamma^{j-k}, & j\geq k\\
         0, & j < k\\
     \end{cases},
 \end{equation}
where $\theta$ is a real parameter vector and $\gamma$ is a positive scalar. Different retention heads have different, fixed $\gamma$ values to capture different types of contextual information. Consequently, retention heads are followed by group, not layer, normalization. RetNets train in parallel form but they perform retention recurrently during inference. Taking $Q$, $K$, and $V$ as in \cref{eqn:parallel_retention}, retention can be expressed as a recurrence on the rows of these three matrices:
\begin{align*}
    & S_t = \gamma S_{t-1} + (K_n)^TV_t, \; S_0 = 0\\
    & \operatorname{Retention}(X_t) = Q_tS_t
    \numberthis
    \label{eqn:recurrent_retention}
\end{align*}

Thus RetNets perform inference in $O(n_\text{seq})$ time, unlike the transformer's $O(n_\text{seq}^2)$. For a RetNet with $2N$ total parameters, $n_b$ decoder blocks, retention dimension $d_\text{attn}$, and sequence length $n_\text{seq}$, the parallel RetNet requires $2N + 4n_\text{b}n_\text{sec}d_\text{attn}$ operations per token---same as transformer---to perform a forward pass. Recurrent RetNet, in contrast, requires $2N + 5n_\text{b}(d_\text{attn})^2$ operations per token \cite{knitter2024retnet_nqs}. While RetNet contains a few more parameters than an equivalent transformer, these results indicate that a recurrent RetNet scales more favorably for problems where $n_\text{seq}>1.75d_\text{attn}$. Though fundamentally less expressive than transformers, RetNets have shown comparable performance on key language modeling baselines \cite{sun2023retnet}. Since NQS training incorporates output from multiple training and inference forward passes, its computational cost is dominated by these forward passes: the dual form of the RetNet makes it useful for improving compute efficiency of NQS, and RetNet ansatze are amenable to solving electronic ground state problems \cite{knitter2024retnet_nqs}.

\subsection{Deriving FLOP Estimates for MADE, RetNet, and Transformer Ans\"atze}
\label{sec:appendix_flop_count}

Following the principles of estimation outlined in \cref{sec:nqs_flop_count}, we now derive the different total FLOP count estimates shown in \cref{tab:flop_count_formulas}. The MADE ansatz is simple to estimate, since it requires a full forward pass through the entire input sequence at all stages of training. Its FLOP estimate is given by

\begin{align*}
    \operatorname{FLOP}_\text{M} &\approx 3N_\text{mod}T\paren{\sum_{m=0}^{\lfloor\log_4 B\rfloor - 1}4^m + B\paren{\frac n 2 - \lfloor\log_4 B\rfloor}} +  BT\paren{(M+3)(3N_\text{mod} + 2N_\text{ph})}\\
    &\approx 3N_\text{mod}T\paren{\frac B 3 + B\paren{\frac n 2 - \log B}} +  BT\paren{(3M+9)(N_\text{mod} + 2(M+3)N_\text{ph})}\\
    &\approx BT\paren{\paren{\frac{3n+2}{2} + 3M + 9 - 3\log B}N_\text{mod} + 2(M+3)N_\text{ph}}.
    \numberthis
    \label{eqn:made_flop}
\end{align*}

We now proceed with RetNets, as in contrast with transformers, calculating the total FLOP estimate remains relatively straightforward: we use the recurrent form of RetNet for \cref{eqn:sample_flop} and the $M$ passes in \cref{eqn:loss_flop}, and the parallel form for the rest, which gives us

\begin{align*}
    \operatorname{FLOP}_\text{R} &\approx BT\paren{\frac 1 3  - \log(B) + \frac{\paren{M+1}n}2}F_\text{rec} + BT\paren{\paren{M+3}F_\text{ph} + \frac {3n} 2 F_\text{par}}\\
    &\approx BT\paren{\paren{2N_\text{mod} + 5d_\text{m}^2n_\text{b}}\paren{\frac{\paren{M+1}n}2 - \log(B)}}\\
    &\;\;\;\;+ BT\paren{\paren{2\paren{M+3}N_\text{ph} + 3n \paren{N_\text{mod} + nd_\text{m}n_\text{b}}}}\\
    &\approx BT\paren{\paren{(M+4)n - 2\log B}N_\text{mod}+2(M+3)N_\text{ph}}\\
    &\;\;\;\; + BT\paren{n_\text{b}d_\text{m}\paren{2.5d_\text{m}\paren{\paren{M+1}n - 2\log B} + 3n^2}}.
    \numberthis
    \label{eqn:retnet_flop}
\end{align*}

Estimating the FLOPs of the transformer ansatz requires consideration of a few different factors. The contribution of the phase network remains unchanged, but unlike RetNets, $F_\text{mod}$ implicitly depends on $n$ during sampling, though we may mitigate this effect to an extent by excluding dummy tokens from forward passes. In this case, we must make adjustments to the expression of \cref{eqn:sample_flop} and get

\begin{align*}
   \operatorname{FLOP}_\text{S(T)} &\approx T\paren{\sum_{m=0}^{\lfloor\log_4 B\rfloor - 1}\!\!\!\! (m+1)4^m\paren{2N_\text{mod} + 4(m+1)n_\text{b}d_\text{m}} + B\!\!\!\!\!\!\!\!\!\!\sum_{m=\lfloor \log_4 B\rfloor + 1}^{n/2}\!\!\!\!\!\!\!\!\!\!\paren{2N_\text{mod} + 4mn_\text{b}d_\text{m}}}\\
   &\approx T\paren{2N_\text{mod}\!\!\!\!\!\!\sum_{m=0}^{\lfloor\log_4 B\rfloor - 1}\!\!\!\!\!\! (m+1)4^m + n_\text{b}d_\text{m}\!\!\!\!\sum_{m=1}^{\lfloor\log_4 B\rfloor}\!\!\!\! m^24^{m}}\\
   &\;\;\;\; + T\paren{BN_\text{mod}\paren{n - 2\log B} + 4Bn_\text{b}d_\text{m}\!\!\!\!\!\!\!\!\!\!\sum_{m=\lfloor \log_4 B\rfloor + 1}^{n/2}\!\!\!\!\!\!\!\!\!\!m\;\;\;}\\
   %&\approx BT\paren{\frac 2 9N_\text{mod}\paren{3\log B - 1} + \frac{4n_\text{b}d_\text{m}}{27}\paren{9\paren{\log B}^2- 6\log B+ 5} + N_\text{mod}\paren{n - \log B^2} + 2n_\text{b}d_\text{m}\paren{\frac {n^2} 4 - \paren{\log B}^2}}\\
   &\approx BT\paren{\paren{n - \frac 4 3\log B - \frac 2 9}N_\text{mod} + n_\text{b}d_\text{m}\paren{\frac{n^2}{2} + \frac{20}{27} - \frac 8 9 \log B  - \frac 2 3\paren{\log B}^2}}.
    \numberthis
    \label{eqn:transformer_sample_flop}
\end{align*}

Estimating the loss count gives

\begin{equation}
   \operatorname{FLOP}_\text{L(T)} \approx BT\paren{(M+3)nN_\text{mod}+ 2(M+3)N_\text{ph} + n_\text{b}d_\text{m}(M+3)n^2},
    \label{eqn:transformer_loss_flop}
\end{equation}

and through combining these two estimates, we obtain the transformer FLOP count estimate

\begin{align*}
   \operatorname{FLOP}_\text{T} &\approx BT\paren{\paren{\paren{M+4}n - \frac 4 3 \log B}N_\text{mod} + 2\paren{M+3}N_\text{ph}}\\
   &\;\;\;\;+ BT\paren{n_\text{b}d_\text{m}\paren{\paren{M+3.5}n^2 - \frac 8 9 \log B - \frac 2 3 \paren{\log B}^2}}.
    \numberthis
    \label{eqn:transformer_flop}
\end{align*}

\subsection{Derivation of Simplified FLOP Estimates}
\label{sec:flop_constrained_derivation}

We now describe the simplification process yielding the rougher FLOP estimates given in \cref{tab:revised_flop}. Traditional LLM compute-constrained scaling law analysis presumes an approximate FLOP count of $6ND$ \cite{kaplan2020scaling}, where $N$ is the model parameter count and $D$ is the number of tokens used from training. This estimate is quite rough, based on the presumption that while the single-token forward pass FLOP count of a transformer is technically greater than $2N$, this difference is not significant in practice.

The NQS FLOP estimates in \cref{tab:flop_count_formulas} are far more elaborate, incorporating several additional terms that dependent on problem size---the qubit number $n$, number $M$ of unique bit flip patterns within the terms of the Hamiltonian, which is roughly $O(n^4)$, and the sample batch size $B$. Furthermore, these formulas do not yet explicitly depend on $N$ and $D'$, the variables used in the NQS parametric scaling law curve from \cref{subsec:parameteric_curves}. Accordingly, we must modify the NQS FLOP estimates both to depend on these two variables explicitly and to be more amenable to the simple constrained optimization performed in scaling law analysis. We give a detailed description for modifying the Transformer ansatz FLOP estimate, but those for RetNet and MADE follow similar principles.

Drawing inspiration from the general principles that lead to the $6ND$ estimate in \cite{kaplan2020scaling}, we first remove all remaining non-leading terms in \cref{eqn:transformer_flop}. We also remove all terms dependent on $\log B$, since $B$ is bounded above by the size of the search space: the discussion of search space size in \cref{subsec:parameteric_curves} and a basic asymptotic analysis of binomial coefficients based on Stirling's approximation together indicate that $\log B$ is $O(n)$, hence it is not one of the primary leading terms in the estimate. We then have a reduced estimate
\begin{equation}
    \operatorname{FLOP}_\text{T} \approx BT\paren{nMN_\text{mod} + 2MN_\text{ph} + n_\text{b}d_\text{m}\paren{Mn^2}}.
    \label{eqn:first_reduced_transformer_flop}
\end{equation}
Looking back to \cite{kaplan2020scaling}, under the model constraints enforced in \cref{sec:experimental_setup}, $N_\text{mod}\approx 12 n_\text{b}d_{\text{m}}^2$, simplifying \cref{eqn:first_reduced_transformer_flop} further to \begin{equation}
    \operatorname{FLOP}_\text{T} \approx BT\paren{M\paren{n + \frac{n^2}{12d_\text{m}}}N_\text{mod} + 2MN_\text{ph}}.
\end{equation} If we let $S$ equal the size of the search space defined by the given NQS problem, and permit ourselves some further overestimate of the FLOP contribution from the phase network, then our reduced FLOP estimate becomes
\begin{equation}
    \operatorname{FLOP}_\text{T} \approx M\paren{n + \frac{n^2}{12d_\text{m}}}SD'N.
\end{equation}
A similar simplification process results in the other entries of \cref{tab:revised_flop}.

\subsection{Brief Overview of Chinchilla Parametric Curve Fitting}
\label{sec:appendix_chinchilla_curve}

A crucial aspect of assessing a model's performance with the Chinchilla scaling law analysis \cite{chinchilla} involves fitting a parametric curve to a collection of data points representing separate training runs of the model: the curve predicts the final training loss value $L$ attained by each instance as a function of model size $N$ and number of tokens $D$ processed by the model over the course of training. The contribution of each variable to the final loss is given via a power law functional form:

\begin{equation}
    L(N, D) = A_0 + \frac{A_1}{N^{\alpha_1}} + \frac{A_2}{D^{\alpha_2}},
    \label{eqn:chinchilla_curve}
\end{equation}
with the final curve being determined through minimizing a Huber loss, taken with $\delta=10^{-3}$, applied to the logarithms of the power law and target loss values and averaged across the entire dataset: \begin{equation}
    \min_{A_i,\alpha_i} \frac 1 R \sum_{j\in R} \operatorname{Huber}_\delta\paren{\log\paren{A_0 + \frac{A_1}{N_j^{\alpha_1}} + \frac{A_2}{D_j^{\alpha_2}}}-\log L(N_j,D_j)}
\end{equation}

Though the exact values of these regression parameters have been disputed \cite{besiroglu2024chinchilla} since the unveiling of the Chinchilla laws, the laws themselves remain a crucial foundation for scaling analysis in LLMs.

\subsection{MADE and RetNet Result Discussion}
\label{sec:appendix_heatmaps}

\begin{figure}
    \centering
    \includegraphics[width=0.95\linewidth]{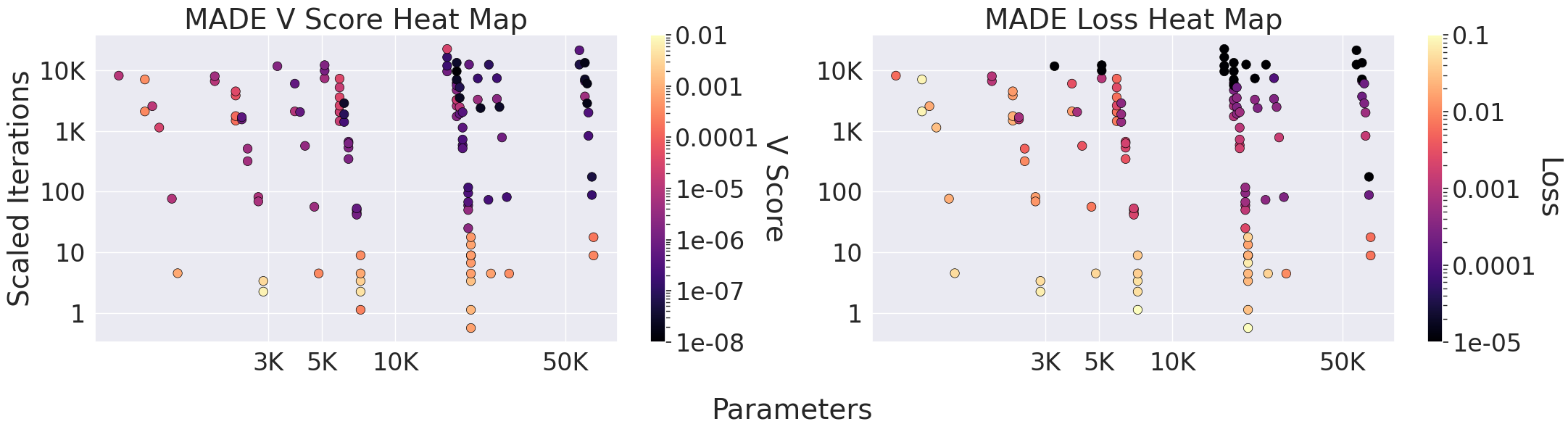}
    \caption{Heat maps showing V-score and absolute error for the MADE ansatz. We note that data points are far less stratified than in \cref{fig:transformer_heatmaps}}
    \label{fig:made_heatmaps}
\end{figure}

We show in \cref{fig:made_heatmaps} the equivalent heat maps for the MADE ansatz to those of the transformer from \cref{fig:transformer_heatmaps}. Observing that these data points are less stratified with respect to model size as can be seen for tranformer and RetNet, we emphasize here that the exact values of $N$ come from both problem size and hidden layer dimension. This observation is technically true for all models discussed in this paper, but the effect is far less visible for the other models, which do not solely consists of feedforward components. The number of hidden layers was deliberately kept fixed and shallow to avoid introducing further sources of error as a result of the specified training hyperparameter configuration being insufficient to handle deeper networks. We do this during testing not only for both networks inside the MADE ansatz, but also for the phase networks, which are feedforward networks, of the RetNet and transformer ansatze.

\begin{figure}
    \centering
    \includegraphics[width=0.95\linewidth]{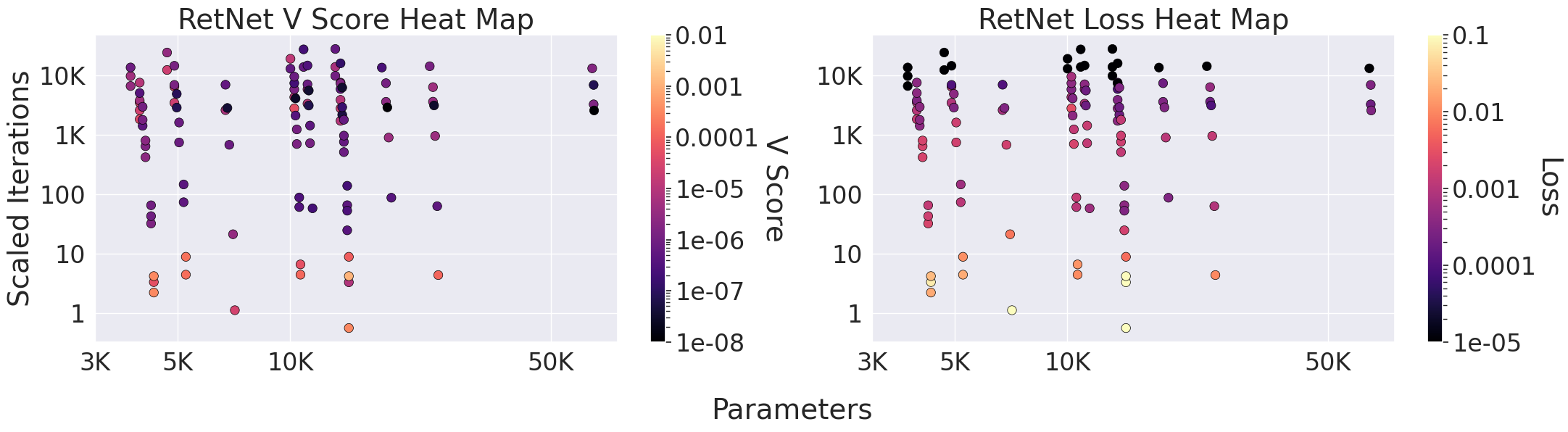}
    \caption{Heat maps showing V-score and absolute error values for the RetNet-based ansatz. We do note that the distribution of data points is more directly similar to \cref{fig:transformer_heatmaps} than to what is shown in \cref{fig:made_heatmaps}.}
    \label{fig:retnet_heatmaps}
\end{figure}

In contrast, the data distribution shown in \cref{fig:retnet_heatmaps} for the RetNet ansatz is almost identical to what can be seen in \cref{fig:transformer_heatmaps}, owing to the fact that RetNets and transformers are so architecturally similar. Of course, the exact V scores and absolute errors attained differ between the two models, but the general trends here corroborate the conclusion from \cite{knitter2024retnet_nqs} that the two models also train quite similarly as NQS ansatze. We can make a few additional general qualitative observations about the observed data. For instance, it appears that, on average, a given combination of independent variables $N$ and $D'$ will yield a lower V-score with a RetNet ansatz than with a transformer, but at the same time, the transformer appears more likely to produce a lower absolute error than the RetNet. A similar comparison between the scaling performance of these architectures and the MADE ansatz is more difficult to make since MADE is a fundamentally different model, but one can observe in the data that RetNet and transformer ansatze show better performance, in terms of both V-score and absolute error, than the MADE ansatz using lower numbers of parameters.

\subsection{Analysis of Compute-Constrained Scaling}
\label{sec:constrained_optimization}

Here we give, in moderate detail, the derivation of the efficient frontier obtained in \cite{chinchilla}, slightly modified for our purposes. Each of the FLOP estimates in \cref{tab:revised_flop} is of the form $kD'N$, for some $k$ specific to the choice of ansatz and the exact problem size. This allows for a similar constrained optimization to \cite{chinchilla}, allowing us to determine from the empirically obtained parametric curve constants the optimal training configuration relative to a fixed compute cost. Both parametric curves in \cref{eqn:vscore_curve} take the form \begin{equation}
    L(N,D') = A_0 + \frac{A_1}{N^{\alpha_1}} + \frac{A_2}{D'^{\alpha_2}}
\end{equation} for some loss function $L$. The method of Lagrange multipliers then says that for a fixed compute budget $C$, we must solve the system of equations \begin{align*}
    \alpha_1 A_1N^{\alpha_1 - 1} &= \lambda kD'\\
    \alpha_2 A_2D'^{\alpha_2 -1} &= \lambda kN\\
    kD'N &= C.
\end{align*} Substituting $D' = \frac{C}{kN}$ into the other two equations gives \begin{align*}
    \alpha_1 A_1 N^{\alpha_1} &= \lambda C\\
    \alpha_2 A_2 C^{\alpha_2 - 1} &= \lambda k^{\alpha_2}N^{\alpha_2},\\
\end{align*} and substituting the first equation into the second then gives \begin{equation}
    \alpha_2 A_2 C^{\alpha_2} = \alpha_1 A_1 k^{\alpha_2} N^{\alpha_1 + \alpha_2},\text{ hence }N = \paren{\frac{\alpha_2 A_2}{\alpha_1 A_1}}^{\frac 1{\alpha_1 + \alpha_2}}\paren{\frac C k}^{\frac{\alpha_2}{\alpha_1 + \alpha _2}}
\end{equation}
Likewise, substituting $N=\frac{C}{kD'}$ into the second equation gives $\alpha_2 A_2 D'^{\alpha_2} = \lambda C$, so \begin{equation}
    D'= \paren{\frac{\alpha_1 A_1}{\alpha_2 A_2}N^{\alpha_1}}^{\frac{1}{\alpha_2}} = \paren{\frac{\alpha_1 A_1}{\alpha_2 A_2}}^{\frac {1}{\alpha_1 + \alpha_2}}\paren{\frac C k}^{\frac{\alpha_1}{\alpha_1 + \alpha _2}}.
\end{equation} These equations for the compute-optimal $N$ and $D'$ are clear analogs of those found in \cite{chinchilla}. %As a result, further 

\end{document}